\documentclass[11pt]{article}

\usepackage[letterpaper,margin=1in]{geometry}
\usepackage{graphicx}
\usepackage{amsmath}
\usepackage{amssymb}
\usepackage[T1]{fontenc}
\usepackage{newtxtext,newtxmath}
\usepackage[hidelinks]{hyperref}
\date{}

\title{\LARGE \bf
NVExplain: Explaining Time Series Forecasting with Latent Trajectory Analysis and Structure-Preserving Surrogates
}

\author{Muyan Anna Li$^{*}$, Manikandan Ravikiran$^{*}$, Aditi Gautam$^{*}$\\[0.6em]
\small *Equal contribution.\\
\small NVIDIA, USA\\
{\tt\small \{annali, mravikiran, adgautam\}@nvidia.com}
}

\begin{document}

\maketitle
\thispagestyle{empty}
\pagestyle{empty}

\begin{abstract}

Time series forecasting models are widely used in high-stakes settings, yet their predictions remain difficult to interpret because existing post-hoc methods often ignore temporal dependence and fail to provide horizon-specific explanations. We propose a model-agnostic explainability framework that explains forecasting predictions by attributing each forecast horizon to temporally relevant historical lags. The framework models forecasting as a latent trajectory and introduces \emph{semantic flow} to quantify how information evolves across time in the model’s internal representations. By aggregating semantic flow, it constructs a lag--horizon attribution matrix that captures horizon-resolved temporal influence. To improve explainability, we further generate structure-preserving perturbations and fit sparse local surrogate models, producing human-readable and temporally coherent explanations. We evaluate the method using faithfulness and stability diagnostics across multiple benchmark datasets. Results show that the semantic-flow variant achieves competitive or superior faithfulness compared to standard post-hoc baselines, while being substantially more computationally efficient. Stability analysis further demonstrates that the explanations are robust and identifies regimes where interpretation should be applied with caution.

\end{abstract}


\section{Introduction}

Time series forecasting underpins high-stakes decisions in energy, finance, transportation, and industrial monitoring. While modern deep learning and foundation-model forecasters achieve strong performance \cite{moment,tft}, their internal reasoning remains opaque, limiting trust and reliable deployment.

Post-hoc explainability methods such as LIME, SHAP, and Integrated Gradients \cite{lime,shap,ig} have been extended to sequential settings via TimeSHAP \cite{timeshap} and ShapTime \cite{shaptime}. However, forecasting poses unique challenges: naive perturbations can violate temporal continuity and seasonality; multi-step forecasting requires horizon-specific reasoning, since the lags influencing near-term predictions differ from those affecting longer horizons; and most methods operate in input space, missing how information evolves through latent representations over time.

We address these gaps with a model-agnostic framework that reframes forecasting explanation as analyzing semantic evolution in latent space. We introduce \emph{semantic flow} to quantify changes between consecutive latent states as the temporal window advances. The framework builds latent trajectories from rolling context windows, computes semantic flow magnitudes, and aggregates them into a lag-by-horizon attribution matrix that yields horizon-resolved distributions over historical lags. To produce interpretable explanations while preserving temporal structure, we employ structure-preserving perturbations and fit horizon-specific sparse local surrogate models.

Evaluated via faithfulness and stability diagnostics against established baselines, semantic flow achieves competitive or stronger faithfulness across multiple datasets while offering more temporally coherent explanations, underscoring the role of latent stability in explanation reliability.

Unlike TimeSHAP and ShapTime, which estimate input-space importance through perturbation or Shapley-style sequence removal, NVExplain explains forecasting behavior through the geometry of latent temporal evolution---a forecasting-specific decomposition that converts rolling latent-state changes into horizon-resolved lag attributions. Semantic flow provides the latent-dynamics primitive, the lag--horizon matrix converts this primitive into multi-step explanations, and the sparse surrogate stage offers an optional human-readable local approximation.

In summary, our contributions are:
\begin{enumerate}
    \item introducing semantic flow as a latent-space explanation primitive,
    \item proposing a model-agnostic lag-horizon attribution framework for multi-step forecasting,
    \item designing structure-preserving perturbations for time series explanations, and
    \item evaluating explanation quality using forecasting-specific faithfulness and stability diagnostics.
\end{enumerate}
\section{Methodology}

Consider a forecasting model
\begin{equation}
f: \mathbb{R}^{C \times L} \rightarrow \mathbb{R}^{C \times H},
\end{equation}
where $C$ is the number of channels, $L$ is the context length, and $H$ is the forecast horizon. Given an input context window
\begin{equation}
X_t = x_{t-L+1:t} \in \mathbb{R}^{C \times L},
\end{equation}
the model produces a multi-horizon forecast
\begin{equation}
\hat{Y}_t = f(X_t) \in \mathbb{R}^{C \times H}.
\end{equation}

Our goal is to explain how information from the input context $X_t$ influences the multi-horizon forecast $\hat{Y}_t$, without modifying or retraining the model. Rather than directly attributing importance in input space, we approach this problem by analyzing how the model’s internal representations evolve over time. \emph{Specifically, we explain predictions by attributing horizon-specific outputs to historical inputs.} The proposed framework consists of four stages: (i) latent trajectory construction, (ii) semantic flow computation, (iii) lag–horizon attribution, and (iv) structure-preserving local surrogate modeling.

\subsection{Latent Trajectory Construction}

To understand how forecasting models process temporal information, we represent the model as inducing a trajectory in latent space by embedding rolling context windows. Let $\mathrm{embed}(\cdot)$ denote the model’s embedding interface. For each rolling window ending at index $\tau$, we compute
\begin{equation}
Z_{\tau} = \mathrm{embed}\!\left(x_{\tau-L+1:\tau}\right) \in \mathbb{R}^{D},
\end{equation}
where $D$ is the latent dimension. This yields a latent trajectory
\begin{equation}
\mathcal{Z} = \{Z_1, Z_2, \dots, Z_T\},
\end{equation}
where $T$ denotes the number of rolling windows for which the embedding is computed, so the trajectory contains one latent state per temporal window. To ensure continuity beyond the observed context, the trajectory is extended using the model’s forecast, allowing analysis over both historical and predicted segments. 
The embedding interface may be instantiated in different ways depending on the deployed forecaster. For encoder-based or encoder-decoder models, it can correspond to the encoder output, pooled hidden state, or penultimate forecasting representation. For decoder-only or API-style systems that do not expose stable internal states, semantic-flow attribution should be interpreted as a grey-box component rather than a strictly black-box one; in such cases, the structure-preserving surrogate stage can still be applied using prediction queries alone, while latent-flow analysis requires either exposed embeddings or a stable proxy representation. Thus, NVExplain is model-agnostic with respect to architecture and training procedure, but its latent-flow variant assumes access to a representation interface.

\subsection{Semantic Flow Magnitudes}

Given a latent trajectory, we aim to quantify how information evolves as the temporal window advances. To this end, we measure changes between consecutive latent states to capture the dynamics of the model’s internal representation. The semantic flow magnitude at step $\tau$ is defined as
\begin{equation}
m_{\tau} = \|Z_{\tau+1} - Z_{\tau}\|_2, \qquad \tau = 1, \dots, T-1.
\end{equation}
This quantity captures the rate of semantic change in the model’s latent state, indicating how strongly new information influences the representation at each step. To reduce sensitivity to noise, the latent trajectory can optionally be smoothed using an exponential moving average:
\begin{equation}
\tilde{Z}_{\tau} = \alpha \tilde{Z}_{\tau-1} + (1-\alpha) Z_{\tau}.
\end{equation}

\subsection{Lag--Horizon Attribution Matrix}

Forecasting requires understanding how past information contributes differently across prediction horizons. To achieve this, we construct lag--horizon attribution distributions that quantify how each historical lag contributes to each forecast step. To this end, we aggregate semantic flow signals over temporal history to construct horizon-specific attribution distributions. Let $K$ denote the number of lags and $H$ the forecast horizon, and let $t$ denote the window index of the prediction being explained. We define a lag--horizon score matrix as
\begin{equation}
S(j,h) = \sum_{a=0}^{j-1} m_{t-1-a} \, w_h(a),
\end{equation}
where $a$ denotes the age of a transition (with $a{=}0$ being the most recent transition before $t$) and $w_h(a)$ is a horizon-dependent kernel. We use an exponential kernel:
\begin{equation}
w_h(a) = \exp\!\left(-\frac{a}{s_h}\right),
\end{equation}
where $s_h$ increases with horizon, reflecting a decay in temporal relevance and allowing longer-term predictions to incorporate broader history.

The final attribution matrix is obtained via a horizon-wise softmax over lags
\begin{equation}
A(j,h)=
\frac{\exp\!\big(S(j,h)/T_{\mathrm{sm}}\big)}
{\sum_{j'}\exp\!\big(S(j',h)/T_{\mathrm{sm}}\big)},
\end{equation}
where $T_{\mathrm{sm}}>0$ is a softmax temperature controlling attribution sharpness (distinct from the trajectory length $T$ and from the window index $\tau$ used above). Each column of $A$ defines a distribution over lags for a fixed horizon, enabling horizon-resolved temporal explanations.

\subsection{Structure-Preserving Local Surrogates}

While the attribution matrix provides a compact global explanation, we further obtain explainable local summaries by fitting sparse surrogate models using temporally coherent perturbations. These perturbed contexts are generated via block-bootstrap resampling combined with Fourier-domain amplitude perturbations, preserving continuity and seasonal structure, and each perturbed sample is represented as
\begin{equation}
u^{(i)} = \mathrm{vec}\!\left(X^{(i)}_{:,\,L-K+1:L}\right).
\end{equation}

For each channel and horizon, we fit a sparse linear surrogate:
\begin{equation}
g_{c,h}(u) = \beta_{0,c,h} + \beta_{c,h}^{\top} u,
\end{equation}
using locality-weighted $\ell_1$ regularization:
\begin{equation}
\min_{\beta_{0,c,h},\,\beta_{c,h}} \sum_{i} w_i
\left(
g_{c,h}(u^{(i)}) - \hat{Y}^{(i)}_{c,h}
\right)^2
+ \lambda \|\beta_{c,h}\|_1.
\end{equation}

Weights are computed in latent space:
\begin{equation}
w_i = \exp\!\left(-\gamma \|Z^{(i)} - Z^{(0)}\|_2^2\right).
\end{equation}

This yields explainable and locally faithful approximations of the model’s forecasting behavior.

Overall, the proposed framework produces three complementary outputs: 
(i) a lag--horizon attribution matrix that captures temporal influence across prediction horizons, 
(ii) semantic flow statistics that characterize latent dynamics, and 
(iii) optional surrogate coefficients that provide explainable local approximations. 
Together, these outputs enable both quantitative and qualitative analysis of forecasting behavior.

\section{Experimental Setup}

\subsection{Model, Datasets, and Parameters}

We evaluate the proposed framework on the MOMENT forecasting model \cite{moment}, treated as a black-box predictor with an accessible embedding interface. The context length is $L=512$, the model horizon is $72$, and explanations are computed over the most recent $K=128$ lags for a forecast horizon of $H=72$. Experiments are conducted on four datasets from the Nixtla long-horizon benchmark collection \cite{datasetsforecast}: ETTh1, Exchange, ILI, and Weather, spanning diverse domains including industrial monitoring, finance, epidemiology, and meteorology. We compare against four baselines: Random, which assigns lag importances drawn uniformly at random and serves as a sanity-check lower bound; Attention \cite{attention}, which uses normalized attention weights; Integrated Gradients (IntGrad) \cite{ig}; and TimeSHAP \cite{timeshap}. We further evaluate two variants of the proposed method: (i) semantic-flow attribution alone, and (ii) the full pipeline with surrogate-based explanations.

\subsection{Evaluation Metrics}

We evaluate the proposed framework along two complementary axes: \emph{faithfulness}, measuring whether explanations capture causal temporal influence, and \emph{stability}, assessing the reliability of latent representations underlying the explanations.

\textbf{Faithfulness.} We use a lag-replacement-based metric. For each horizon $h$, let $\Delta^{\mathrm{abs}}_h = |\hat{y}^{\mathrm{top}}_h - \hat{y}_h|$ denote the absolute forecast change after replacing the top-$k$ lags with the channel-wise dataset mean, and define the relative effect as
\begin{equation}
\Delta^{\mathrm{rel}}_h =
\frac{\Delta^{\mathrm{abs}}_h}
{|\hat{y}_h| + \epsilon}.
\end{equation}
A horizon is counted as a pass if $\Delta^{\mathrm{rel}}_h \geq \alpha$ and $\Delta^{\mathrm{abs}}_h \geq \gamma\,\Delta^{\mathrm{abs},\mathrm{ctrl}}_h$, where $\Delta^{\mathrm{abs},\mathrm{ctrl}}_h$ is the absolute change from random and bottom-ranked lag removals, $\alpha$ is a minimum effect threshold, and $\gamma>1$ is a control margin. The overall faithfulness is computed as
\begin{equation}
\mathrm{FPR} = \frac{1}{H}\sum_{h=1}^{H}\mathbf{1}\{\text{horizon } h \text{ passes}\}.
\end{equation}

\textbf{AOPC and Gap.} AOPC measures cumulative forecast degradation as top-ranked lags are progressively removed \cite{forecast_eval,timeshap}. 
The faithfulness Gap is the difference in AOPC between top- and bottom-ranked lag removal. 
Higher AOPC and larger Gap indicate more faithful identification of consequential lags. Reported values average over horizons.

\textbf{Stability.} Since the method relies on latent trajectories, we assess reliability using two diagnostics. \emph{Embedding stability} measures sensitivity of latent representations to small input perturbations, defined as
\begin{equation}
R_t =
\frac{\|\mathrm{embed}(X'_t) - \mathrm{embed}(X_t)\|_2}
{\|X'_t - X_t\|_F + \epsilon}.
\end{equation}
We report summary statistics (mean and $95$th percentile) over perturbation trials. \emph{Trajectory stability} captures smoothness of latent evolution and is summarized using zero-crossing rate, direction flip, and relative jitter averaged over latent dimensions. These diagnostics are particularly relevant in forecasting, where temporally consistent representations are essential for reliable explanations. Lower values indicate more stable and trustworthy explanations.

\section{Preliminary Results}

We evaluate the proposed method using faithfulness metrics (FPR and AOPC) and stability diagnostics, comparing against Random, Attention \cite{attention}, Integrated Gradients \cite{ig}, and TimeSHAP \cite{timeshap}. We report two variants: Ours (flow), using semantic-flow attribution directly, and Ours (full), incorporating surrogate-based explanations.

Table~\ref{tab:benchmark_faithfulness} reports faithfulness pass rate (FPR). The semantic-flow variant matches the best result on ETTh1 ($99.1\%$) and outperforms all baselines on Exchange ($88.9\%$) and Weather ($96.3\%$). On ILI, Integrated Gradients scores highest ($76.1\%$), while semantic flow remains competitive ($67.1\%$) at far lower cost---indicating that latent-state transitions identify causally relevant lags more effectively than input-space attribution.

\begin{table}[t]
\caption{Faithfulness pass rate (FPR, \%) across datasets.}
\label{tab:benchmark_faithfulness}
\centering
\setlength{\tabcolsep}{5pt}
\begin{tabular}{l c c c c}
\hline
Method & ETTh1 & Exchange & ILI & Weather \\
\hline
Random          & 13.9 & 17.3 & 17.0 & 7.4 \\
Attention       & 21.8 & 13.3 & 15.6 & 54.2 \\
IntGrad         & 83.8 & 83.8 & 76.1 & 48.5 \\
TimeSHAP        & 99.1 & 67.6 & 44.4 & 64.8 \\
Ours (flow)     & 99.1 & 88.9 & 67.1 & 96.3 \\
Ours (full)     & 96.8 & 59.9 & 50.6 & 72.1 \\
\hline
\end{tabular}
\end{table}

\begin{figure}[t]
\centering
\includegraphics[width=0.99\columnwidth]{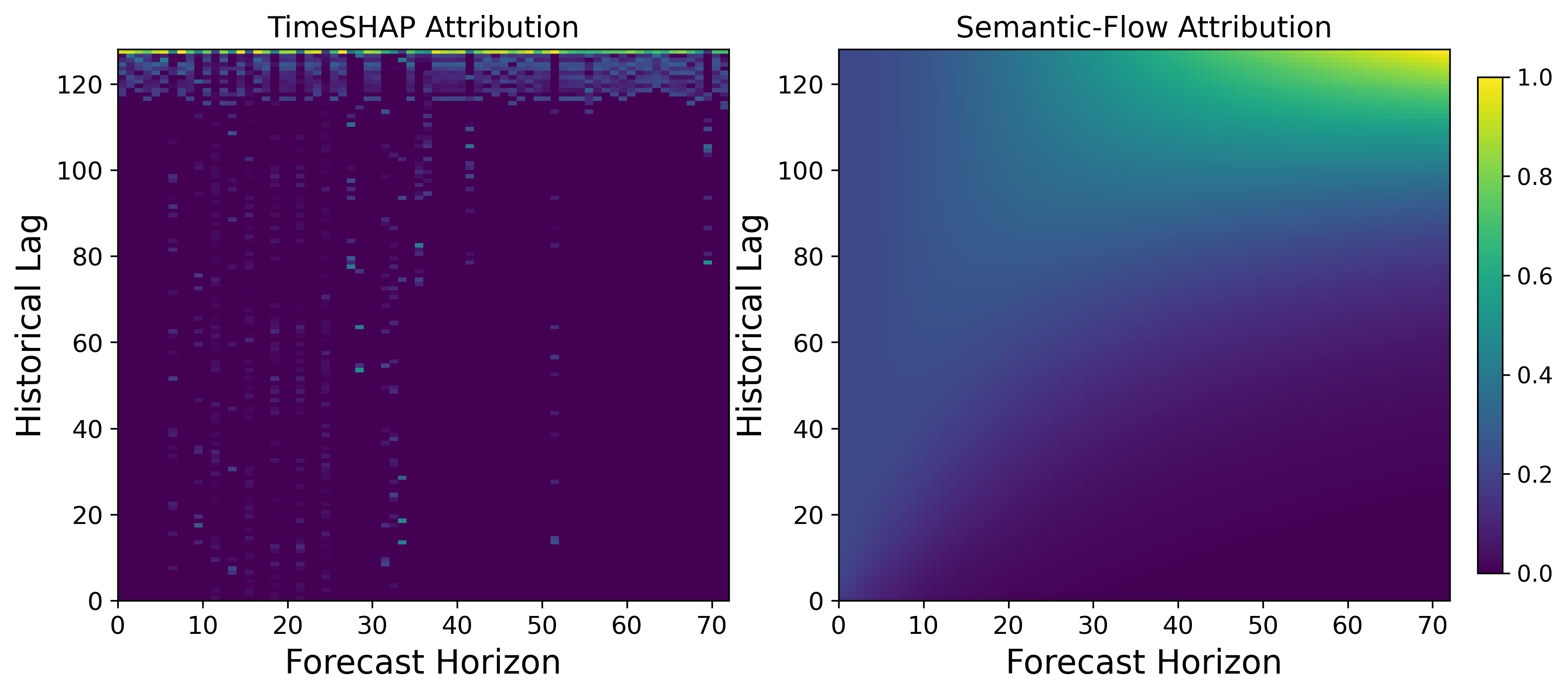}
\caption{Lag-by-horizon attribution on Weather. TimeSHAP (left) is fragmented and noisy ($\mathrm{TV}{=}0.474$); the semantic-flow attribution (right) concentrates on recent lags and decays smoothly across horizons ($\mathrm{TV}{=}0.010$).}
\label{fig:lag_horizon_compare}
\end{figure}

Fig.~\ref{fig:lag_horizon_compare} compares attribution maps from TimeSHAP and semantic flow on Weather.

Table~\ref{tab:benchmark_secondary} shows semantic flow yields the largest forecast degradation (AOPC) and strongest separation from control perturbations on ETTh1, Exchange, and Weather, indicating the identified lags carry substantial predictive signal, while also running significantly faster than Integrated Gradients and TimeSHAP.

\begin{table}[t]
\caption{Secondary metrics: AOPC (higher is better), Gap, and runtime (s).}
\label{tab:benchmark_secondary}
\centering
\begin{tabular}{l l c c c}
\hline
Dataset & Method & AOPC & Gap & Time (s) \\
\hline
ETTh1 & Random      & 0.115 & 0.002 & 0.000 \\
      & Attention   & 0.229 & -0.087 & 0.037 \\
      & IntGrad     & 0.277 & 0.193 & 52.420 \\
      & TimeSHAP    & 0.393 & 0.343 & 19.784 \\
      & Ours (flow) & 0.442 & 0.392 & 1.952 \\
      & Ours (full) & 0.392 & 0.313 & 2.411 \\
\hline
Exchange & Random      & 0.056 & 0.001 & 0.000 \\
         & Attention   & 0.122 & -0.090 & 0.014 \\
         & IntGrad     & 0.173 & 0.134 & 53.918 \\
         & TimeSHAP    & 0.246 & 0.222 & 31.175 \\
         & Ours (flow) & 0.279 & 0.255 & 5.611 \\
         & Ours (full) & 0.183 & 0.150 & 9.244 \\
\hline
ILI & Random      & 0.038 & -0.003 & 0.000 \\
    & Attention   & 0.061 & 0.015 & 0.079 \\
    & IntGrad     & 0.177 & 0.168 & 232.081 \\
    & TimeSHAP    & 0.074 & 0.028 & 159.294 \\
    & Ours (flow) & 0.159 & 0.116 & 10.305 \\
    & Ours (full) & 0.120 & 0.081 & 21.932 \\
\hline
Weather & Random      & 0.026 & 0.000 & 0.000 \\
        & Attention   & 0.092 & -0.005 & 0.098 \\
        & IntGrad     & 0.067 & 0.048 & 258.804 \\
        & TimeSHAP    & 0.071 & 0.060 & 132.732 \\
        & Ours (flow) & 0.160 & 0.145 & 15.133 \\
        & Ours (full) & 0.101 & 0.075 & 71.099 \\
\hline
\end{tabular}
\end{table}

Ours (full) remains strong on ETTh1 but degrades on Exchange, ILI, and Weather, suggesting surrogate models introduce approximation error under highly nonlinear temporal dynamics.

Table~\ref{tab:stability} reports latent-representation stability. Exchange, ILI, and Weather show low embedding sensitivity (below $0.02$), indicating stable trajectories suited to semantic flow analysis. ETTh1 is an outlier with much higher sensitivity, yet still achieves the highest faithfulness, suggesting strong temporal dependence can persist even under unstable latent dynamics.

Beyond faithfulness, attribution maps offer a practical readability signal: fragmented maps scatter importance across unrelated lags, while smoother horizon-wise patterns are easier to interpret, as illustrated by the noisy TimeSHAP map versus the concentrated, smoothly decaying semantic-flow structure on Weather. We treat this as a qualitative diagnostic rather than a substitute for a formal human-subject study, left to future work.

\begin{table}[t]
\caption{Embedding stability (mean and $95$th percentile).}
\label{tab:stability}
\centering
\begin{tabular}{l c c}
\hline
Dataset & Mean & P95 \\
\hline
Weather  & 0.0084 & 0.0101 \\
ILI      & 0.0100 & 0.0128 \\
Exchange & 0.0187 & 0.0237 \\
ETTh1    & 0.3683 & 0.4841 \\
\hline
\end{tabular}
\end{table}

Table~\ref{tab:trajectory} reports trajectory stability: Weather shows the smoothest latent evolution, while ETTh1 shows the highest volatility, indicating smoother dynamics yield more reliable semantic flow estimates.

\begin{table}[t]
\caption{Trajectory stability diagnostics.}
\label{tab:trajectory}
\centering
\begin{tabular}{l c c c}
\hline
Dataset & Zero-Crossing & Direction Flip & Jitter \\
\hline
ETTh1    & 0.2477 & 0.6352 & 0.8182 \\
Exchange & 0.0956 & 0.6590 & 0.5056 \\
ILI      & 0.1091 & 0.5541 & 0.4600 \\
Weather  & 0.0701 & 0.6080 & 0.3533 \\
\hline
\end{tabular}
\end{table}

Together, these results justify the method's explainability: faithfulness metrics confirm identified lags causally impact predictions, stability diagnostics confirm robustness across perturbations and temporal evolution, and horizon-resolved attribution enables localized explanations unavailable to global attribution methods.

Three insights emerge. First, semantic flow drives strong faithfulness across most datasets, showing latent-state dynamics capture meaningful temporal influence. Second, surrogate models trade explainability for fidelity loss in nonlinear settings. Third, faithfulness and stability are complementary---faithfulness reflects causal influence, stability reflects representation robustness. Datasets like Weather, with both high faithfulness and stable latents, offer the most reliable interpretive regime, whereas unstable regimes warrant more cautious use.
\section{Conclusion}

We introduced NVExplain, an architecture-agnostic explainability framework for time series forecasting that attributes each forecast horizon to relevant historical lags through semantic flow in latent space. The method provides horizon-specific temporal explanations, achieves competitive faithfulness with lower computational cost than standard post-hoc baselines, and includes stability diagnostics to assess explanation reliability. The results also clarify a key limitation of the surrogate stage: in highly nonlinear temporal regimes, sparse local surrogates may sacrifice fidelity even when semantic-flow attribution remains faithful. Therefore, we recommend using the full surrogate explanation when local linearity diagnostics are acceptable, and relying on the semantic-flow attribution matrix together with stability diagnostics when the surrogate fit is unreliable. While the latent-flow component assumes access to an embedding or stable representation interface, the structure-preserving surrogate stage can still support prediction-query-based local explanation. Future work will extend the approach to multivariate forecasting, combine lag-level attribution with feature-level explanations, and evaluate interpretability through user-centered studies.

\end{document}